\documentclass[nohyperref,table,xcdraw,dvipsnames]{article}

\usepackage[accepted]{icml2022}

\usepackage{microtype}
\usepackage{graphicx}
\usepackage{booktabs}

\usepackage{bbding}
\usepackage{multirow}
\usepackage{float}
\usepackage{dsfont}
\usepackage{bm}
\usepackage{wrapfig}
\usepackage{rotating}

\usepackage{balance}

\usepackage{amsmath,amsfonts,amssymb}
\usepackage{mathtools}
\usepackage{amsthm}
\usepackage{subfig}
\usepackage{epstopdf}
\usepackage[ruled,vlined,algo2e]{algorithm2e}
\usepackage{mathrsfs}
\usepackage[nocomma]{optidef}
\usepackage[hidelinks]{hyperref}
\usepackage{xspace}

\usepackage{color, colortbl}
\definecolor{LightCyan}{rgb}{0.88,1,1}
\newcommand{\poet}{POET}
\newcommand{\poetlong}{Private Optimal Energy Training}

\icmltitlerunning{\poet{}: Training Neural Networks on Tiny Devices with Integrated Rematerialization and Paging}

\begin{document}

\twocolumn[
\icmltitle{\poet{}: Training Neural Networks on Tiny Devices\\with Integrated Rematerialization and Paging}

\icmlsetsymbol{equal}{*}

\begin{icmlauthorlist}
\icmlauthor{Shishir G. Patil}{ucb}
\icmlauthor{Paras Jain}{ucb}
\icmlauthor{Prabal Dutta}{ucb}
\icmlauthor{Ion Stoica}{ucb}
\icmlauthor{Joseph E. Gonzalez}{ucb}
\end{icmlauthorlist}

\icmlaffiliation{ucb}{University of California Berkeley}

\icmlcorrespondingauthor{Shishir G. Patil}{\texttt{shishirpatil@berkeley.edu}}

\icmlkeywords{Edge Training, Rematerialization, Paging, Machine Learning, ICML}

\vskip 0.4in
]

\begin{abstract}

Fine-tuning models on edge devices like mobile phones would enable privacy-preserving personalization over sensitive data. However, edge training has historically been limited to relatively small models with simple architectures because training is both memory and energy intensive.
We present \poet{}, an algorithm to enable training large neural networks on memory-scarce battery-operated edge devices.
\poet{} jointly optimizes the integrated search search spaces of \textit{rematerialization} and \textit{paging}, two algorithms to reduce the memory consumption of backpropagation.
Given a memory budget and a run-time constraint, we formulate a mixed-integer linear program (MILP) for energy-optimal training.
Our approach enables training significantly larger models on embedded devices while reducing energy consumption while not modifying mathematical correctness of backpropagation.
We demonstrate that it is possible to fine-tune both ResNet-18 and BERT within the memory constraints of a Cortex-M class embedded device while outperforming current edge training methods in energy efficiency. POET is an open-source project
available at 
\url{https://github.com/ShishirPatil/poet}
\end{abstract}

\printAffiliationsAndNotice{}

\section{Introduction}
\label{sec:intro}
Deep learning models are widely deployed for inference on edge devices like smartphones and embedded platforms. 
In contrast, training is still predominantly done on large cloud servers with high-throughput accelerators such as GPUs. 
The centralized cloud training model requires transmitting sensitive data from edge devices to the cloud such as photos and keystrokes, thereby sacrificing user privacy and incurring additional data movement costs.

To enable users to personalize their models without relinquishing privacy, on-device training methods such as federated learning~\cite{federated-smith} perform local training updates without the need to consolidate data to the cloud.
These methods have been widely deployed to personalize keyboard suggestions in Google Gboard~\cite{google-keyboard-federated} and to improve Automatic Speech Recognition (ASR) on iPhones~\cite{apple-ml}.  
\begin{figure*}[t]
    \centering
     \includegraphics[width=\linewidth]{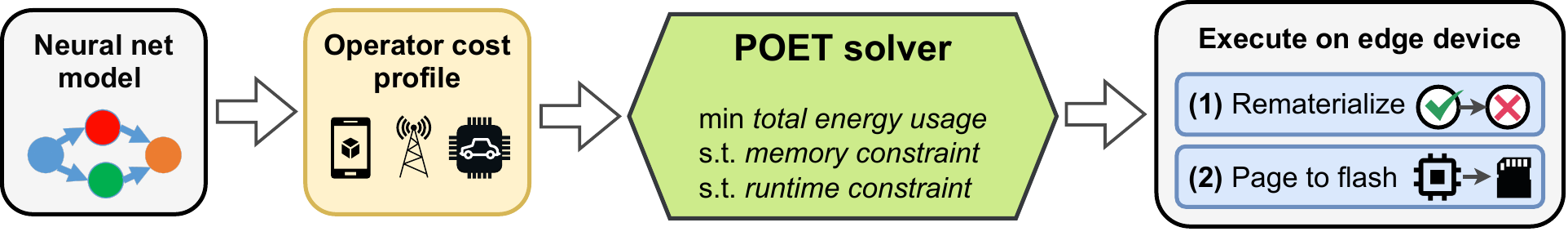}
    \caption{\poet{} optimizes state-of-the-art ML models for training on Edge devices. Operators of the ML model are profiled on target edge device to obtain fine-grained profiles. \poet{} adopts an integrated integrated rematerialization and paging to produce an energy-optimal training schedule.}
    \label{fig:poet-overview}
\end{figure*}

At the same time, current on-device training methods cannot support training modern architectures and large models. For example, Google Gboard fine-tunes a simple logistic regression model. Training larger models on edge devices is infeasible primarily due to the limited device memory which cannot store activations for backpropagation. A single training iteration for ResNet-50~\cite{resnet} requires 200$\times$ more memory than inference.

Prior work has proposed strategies including paging to auxiliary memory~\cite{capuchin} and rematerialization~\cite{chen2016b,checkmate,dtr} to reduce the memory footprint of training in the cloud. However, these methods result in a significant increase in total energy consumption. 
The data transfers associated with paging methods often require more energy than recomputing the data. 
Alternatively, rematerialization increases energy consumption at a rate of $O(n^2)$ as the memory budget shrinks.

In this work, we show that \emph{paging and rematerialization are highly complementary}. By carefully rematerializing cheap operations while paging results of expensive operations to auxiliary memory such as a flash or an SD card, we can scale effective memory capacity with minimal energy overhead. By combining these two methods, we demonstrate it is possible to train  models like BERT on mobile-class edge devices. By framing edge training as an optimization problem, we discover optimal schedules with provably minimal energy consumption at a given memory budget. While the focus of this paper is edge deployemnts, the energy objective is increasingly becoming relevant even for cloud deployments~\cite{patterson2022carbon}.

We present \poet{}~(\poetlong{}), an algorithm for energy-optimal training of modern neural networks on memory-constrained edge devices (Fig~\ref{fig:poet-overview}).  Given that it is prohibitively expensive to cache all activation tensors for backpropagation, \poet{} optimally pages and rematerializes activations, thereby reducing memory consumption by up to 2x. We reformulate the edge training problem as an integer linear program (ILP) and find it is solved to optimality in under ten minutes by commodity solvers.

For models deployed on real-world edge devices, training happens when the edge device is relatively idle and spare compute cycles are available. For example, Google Gboard schedules model updates when the phone is put to charge. Hence \poet{} also incorporates a hard training constraint. Given a memory constraint and the number of training epochs, \poet{} generates solutions that also satisfy the given training deadline. \poet{} transparently develops a comprehensive cost model by profiling the target hardware with the target network's operators. Finally, \poet{} is mathematically value preserving (\emph{i.e,} it makes no approximations), and it works for existing architectures out-of-the-box.

The novel contributions of this work include: 

\begin{enumerate}
    \item A formulation of an integer linear program to find the \textbf{energy-optimal} schedule to train modern deep neural networks \textbf{with a) memory and b) runtime as hard constraints}.
    \item A unified algorithm for hybrid activation recomputation and paging.
    \item The first demonstration of how to train ResNet-18 and BERT on tiny Cortex M class devices with memory and timing constraints. 
\end{enumerate}

\section{Related Work}
\label{sec:related}

Scarcity of compute and memory is one of the largest constraint for machine learning on edge devices. Large models with state-of-the-art performance have largely been exorbitantly expensive for edge devices. The research community has predominantly focused on addressing \textit{inference} on edge devices via methods like efficient DNN architecture~\cite{squeezenet, efficientnetv2}, quantization~\cite{hawq} or pruning~\cite{state_of_pruning}.

Instead, we aim to make \textit{training} large neural networks feasible on tiny edge devices. While compute is the limiting resource for inference on the edge, limited memory capacity constraints prevent training large models on the edge. Training via vanilla backpropagation requires caching the output of all intermediate layers (activations). We categorize methods to reduce memory usage of training as activation (1) compression, (2) rematerialization, and (3) paging. We then discuss prior work in energy-efficient training.

\textbf{Activation quantization:}~~~
\citet{chen2021actnn}, \citet{park-quantization}, and others have proposed techniques to quantize activations while performing full-precision multiply-accumulates (MACs). However, these techniques compromise accuracy and correctness. Moreover, poor hardware support for quantized operations under 8 bits limits the practical savings of these techniques. We do not consider methods for pruning during training like \citet{lotterytickethypothesis} as they do not reduce the size of activations.

\textbf{Rematerialization:}~~~
Rematerialization discards activations in the forward pass and recomputes those values during gradient calculation. \citet{chen2016b} proposed a simple and widely used algorithm for rematerialization where every $O(\sqrt{n})$ layer is retained for the backward pass. \citet{griewank}~propose an optimal algorithm for rematerialization on unit-cost linear auto-diff graphs. However, they force the strong assumption that models have uniform compute requirements across layers.
Checkmate~\cite{checkmate} identifies the optimal rematerialization schedule for arbitrary static graphs. \citet{monet} extends Checkmate with operator implementation selection, but this is orthogonal to our work's scheduling problem. Dynamic Tensor Rematerialization (DTR)~\cite{dtr} finds an approximation of Checkmate that is near-optimal for common computer-vision models. Our work addresses the following limitations of Checkmate: (1) Checkmate does not consider energy nor latency as a constraint and (2) Checkmate does not page activations to secondary memory. \poet{} is the first work that demonstrates provably optimal integrated paging and rematerialization.

\textbf{Paging:}~~~
\citet{swapadvisor} and \citet{deepspeedoffload} page activations off a memory-scarce GPU to the CPU when out of memory. However, we find paging is very energy-intensive and is often less efficient than rematerialization.
Capuchin~\cite{capuchin} uses the Memory Saving Per Second (MSPS) heuristic to decide what to page. Only if paging is insufficient will Capuchin rematerialize activations thereby making it sub-optimal as demonstrated in Sec~\ref{sec:experiments}. 
POFO \cite{beaumont2021}  formulates finding finding the optimal sequence combining rematerialization and paging as a dynamic programming problem. POFO  makes many assumptions that limit generality: POFO only supports chain (linear) model graphs while we support arbitrary graphs such as BERT (Fig~\ref{fig:pareto_flops}). POFO limits layers to a single rematerialization or page operation while \poet{} can remat/page layers repeatedly. POFO forces all page-out operations to occur prior to calculating the loss while we have no such restriction.
And finally, while POFO assumes paging is asynchronous (e.g., CUDA) but this is not universally true for the edge devices we evaluate.
Notice that \poet{} is not only optimizing a different metric (energy vis-a-vis time) but a) adhere's to strict timing guarantees and b) is \emph{provably optimal}.

\textbf{Energy-efficient training:} We are not the first to consider energy-optimal training. Prior work on energy-optimal training for the edge either a) required the design of new architectures~\cite{proxylessnas, efficientnetv2}, or proposed b) new techniques of training by dropping activations, updating only select layers of the network, or c) used a different optimizer~\cite{e2-neurips19}. Compared to these techniques, \poet{} is a) mathematically value preserving (makes no approximations, or modifications), and b) works for existing and new architectures out-of-the-box.  

\begin{table}[t]
\centering

\resizebox{\linewidth}{!}{%
\begin{tabular}{@{}lcccc@{}}
\toprule
\multicolumn{1}{c}{\textbf{Method}}& \begin{tabular}[c]{@{}c@{}}\textbf{General} \\ \textbf{Graphs}\end{tabular} & \begin{tabular}[c]{@{}c@{}}\textbf{Compute} \\ \textbf{Aware}\end{tabular} & \begin{tabular}[c]{@{}c@{}}\textbf{Memory} \\ \textbf{Aware}\end{tabular} & \begin{tabular}[c]{@{}c@{}}\textbf{Power} \\ \textbf{Aware}\end{tabular} \\ \midrule
Checkpoint all (PyTorch) & $\surd$              & $\times$          & $\times$            & $\times$           \\
\citet{griewank}  & $\times$              & $\times$          & $\times$            & $\times$           \\
\citet{chen2016b}~$\sqrt{n}$& $\times$              & $\times$          & $\times$            & $\times$           \\
\citet{chen2016b}~greedy & $\times$              & $\times$          & $\sim$          & $\times$           \\ \midrule
Checkmate~\cite{checkmate}  & $\surd$              & $\surd$          & $\surd$            & $\times$           \\  
POFO ~\cite{beaumont2021} & $\times$              & $\surd$          & $\surd$          & $\times$           \\
DTR~\cite{dtr} & $\surd$              & $\surd$          & $\surd$          & $\times$           \\  
  \midrule
	
\rowcolor{LightCyan}
\poet{} (ours)                  & $\surd$              & $\surd$          & $\surd$            & $\surd$           \\ \bottomrule
\end{tabular}
}
\caption{Comparison of baseline methods under power, compute, memory and generality metrics. \poet{} satisfies all criteria, enabling end-to-end training on the edge.}
\end{table}

\section{Background}
\label{sec:motivation}

\begin{figure}[t]
    \centering
     \includegraphics[width=0.7\linewidth]{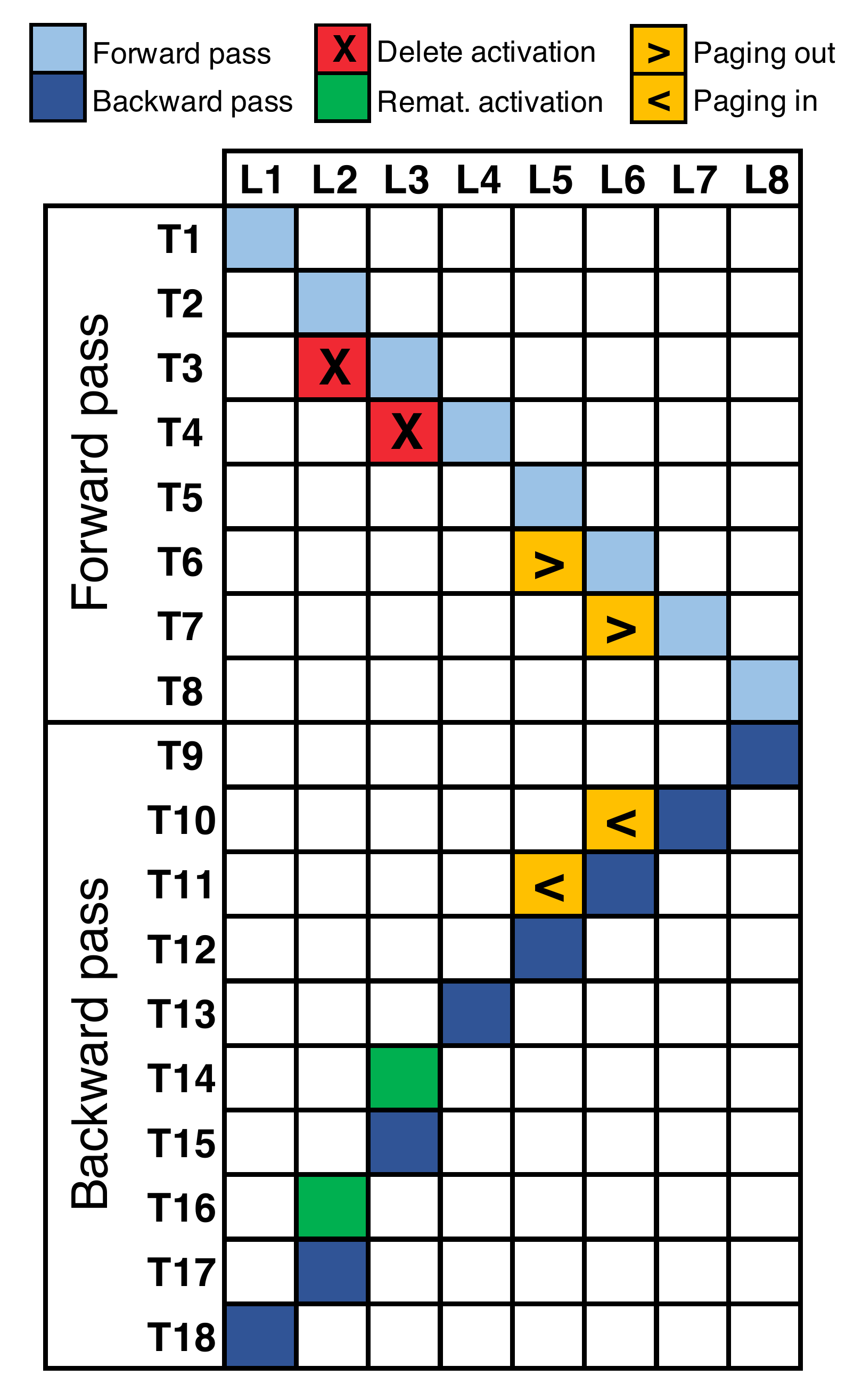}
    \caption{Rematerialization and paging are complementary. This plot visualizes the execution schedule for an eight layer neural network. We represent logical timesteps in increasing order on the y-axis while different layers are represented by the x-axis. Layers 2 and 3 are cheap-to-compute operators and therefore can be rematerialized at low cost. However, layers 5 and 6 are compute intensive so it is more energy-efficient to page them to secondary flash storage.}
    \label{fig:illustrated_example}
\end{figure}

A growing demand exists for edge machine learning applications for greater autonomy. In response, the community has developed systems to enable machine learning on edge devices. EdgeML~\cite{edgeml}, CoreML~\cite{coreml} and {TensorFlow Lite}~\cite{tf-lite} from Google are all efforts to meet this demand.

However, each of these efforts is application-specific and proposes new algorithms or optimizations to address the computational and memory requirements for machine learning inference on the edge~\cite{deep-x}. While inference is already commonly deployed on edge devices, training remains ad-hoc and infeasible for large models.

Training on the edge is critical for privacy, cost, and connectivity. 
First, due to privacy concerns,
many edge applications cannot transmit data to the cloud.
Second, the energy consumed by bulk data transmission can significantly reduce battery life~\citep{trickle}. 
Third, applications such as ocean sensing and communication~\citep{fadel-underwater}, and those deployed in farms~\citep{farmbeats} are designed for offline operations -- with no access to the internet.

Our objective of optimizing for energy is a non-trivial contribution. On edge devices, the energy objective can oftentimes conflict with the objective of running to completion. 
For example, on a given platform, rematerializing might consume lower energy, but paging might be quicker. This is because, on edge devices, it is common practise to turn-off/duty-cycle components that are not utilized (e.g., SD card, DMA, etc.)  The energy profile may vary depending on the size of the tensor, and if the PCIe/DMA/SPI/I2C memory bus needs to be activated, etc. 
Exploiting this enables \poet{} to find the most energy-efficient schedule which would not have been possible had we not optimized for energy. 

While definitions differ on which devices are included in ``the edge'' (e.g., mobile phones, routers, gateways, or even self-driving cars). In the context of this paper, the edge refers to mobile phones and microcontrollers (Table~\ref{table:device-spec}). These devices are characterized by limited memory (ranging from KBs to a few GBs) and are commonly battery-powered ($\sim$ few hundred mAh) for real-world deployment. 
Further, in our research we found that it is quite common for these edge devices to be augmented with an off-chip secondary storage such as a flash or an SD card as seen in ~\cite{farmbeats, fadel-underwater, gesturepod}. This presents us with an opportunity to exploit the off-chip memory for paging.

\section{Integrated paging and rematerialization}

Rematerialization and paging are two techniques to lower the memory consumption of large, state-of-the-art ML models. In rematerialization, an activation tensor is deleted as soon as they are no longer needed, most often,  during a forward pass. This frees up precious memory that can be used to store the activations of the following layers. When the deleted tensor is needed again, for example, to compute gradients during backpropagation, it is recomputed from the other dependent activations as dictated by the lineage. Paging, also known as offloading, is a complementary technique to reduce memory. In paging, an activation tensor that  is not immediately needed is paged-out from the primary memory to a secondary memory such as a flash or an SD card. When the tensor is needed again, it is paged back in. 

This is best understood with the representative neural-network training timeline from Figure~\ref{fig:illustrated_example}. Along the X-axis, each cell corresponds to a single layer of an eight-layered, linear, neural-network. The Y-axis represents the logical timesteps over one epoch. An occupied cell indicates that an operation (forward/backward pass computation, rematerialization, or paging) is executed at the corresponding timestep. For example, we can see that the activation for Layer 1 (L1) is computed at the first timestep (T1). At timestep T2 and T3, the activations of L2 and L3 are computed respectively. Suppose layers L2 and L3 happen to be memory-intensive but cheap-to-compute operators, such as non-linearities (tanH, ReLU, etc,) then rematerialization becomes the optimal choice. We can  delete the activations (\{T3, L2\}, \{T4, L3\}) to free up memory, and when these activations are needed during backward propagation we can rematerialize them (\{T14, L3\}, \{T16, L2\}).  

Suppose layers L5 and L6 are compute-intensive operators such as convolutions, dense matrix-multiplication, etc. For such operations, rematerializing the activations would lead to an increase in run-time and energy and is sub-optimal. For these layers, it is optimal to page-out the activation tensor to secondary storage (\{T6,L5\}, \{T7, L6\}), and page-in when they are needed (\{T10,L6\}, \{T11, L5\}).

One major advantage of paging is that depending on how occupied the memory bus is, it can be pipelined to hide latency. This is because modern systems have DMA (Direct Memory Access) which can move the activation tensor from the secondary storage to the primary memory while the compute engine is running in parallel. For example, at timestep T7, we are both paging L6 out and computing L7. However, rematerialization is compute-intensive, cannot be parallelized. This leads to an increase in run-time. For example, we have to dedicate timestep T14 to recompute L3 thereby delaying the rest of the backward pass execution.

\section{\poet{}: \poetlong{}}
\label{sec:poet}
We introduce \poetlong{} (\poet), a graph-level compiler for deep neural networks that rewrites training DAGs for large models to fit within the memory constraints of edge devices while remaining energy-efficient.
\poet{} is hardware-aware and first traces the execution of the forward and backward pass with associate memory allocation requests, runtime, and per-operation memory and energy consumption. This fine-grained profiling for each workload happens only once for a given hardware, is automated, cheap, and provides the most accurate cost model for \poet{}. \poet{} then generates a Mixed Integer Linear Programming (MILP) which can be efficiently solved. The \poet{} optimizer searches for an efficient rematerialization and paging schedule that minimizes end-to-end energy consumption subject to memory constraints. The resulting schedule is then used to generate a new DAG to execute on the edge device. While the MILP is solved on commodity hardware, the generated schedule shipped to the edge device is only a few hundred bytes, making it highly memory efficient.

Rematerialization is most efficient for operations that are cheap-to-compute yet memory-intensive. 
These operations can be recalculated with low energy overhead. 
Paging, however, is best suited to compute-intensive operations where rematerialization would otherwise incur significant energy overhead. \poet{} jointly considers both rematerialization and paging in an integrated search space.

Without a minimum training throughput limit, it is possible that the energy optimal strategy is also far too slow to train in practical applications.
In reality, training needs to run while the device is idle where spare compute cycles are available. For example, Google Android schedules ML model updates when the phone is charging. To maintain high training throughputs, the \poet{} optimizer can maintain a minimum training throughput to ensure that training completes during downtime.

Given a memory budget $\mu_{RAM}$ and a training time budget $\mu_{deadline}$, \poet{} finds an energy optimal schedule by choosing to either a) rematerialize or b) page the tensors to/from secondary storage such as an SD card. Our method scales to complex, realistic architectures and is \emph{hardware-aware} through the use of
microcontroller-specific, profile-based cost models. We build upon the formulation proposed by Checkmate~\cite{checkmate} and adapt it to jointly consider integrated rematerialization and paging, to optimize for an energy objective rather than the runtime, and to implement a minimum throughput constraint.

\begin{algorithm*}[t!]

\begin{argmini}
    {}{\sum_T \left[R \Phi_{compute} + M_{in} \Phi_{pagein} + M_{out} \Phi_{pageout}\right]_T}{}{}
    \addConstraint{R_{t, i} + S^{RAM}_{t, i}}{\geq R_{t, j}\qquad}{\forall t \in V ~~\forall (v_i, v_j) \in E}
    \addConstraint{R_{t-1, i} + S^{RAM}_{t-1, i} + M^{in}_{t-1, i}}{\geq S^{RAM}_{t, i}}{\forall k \in K ~~\forall t \geq 2 ~~\forall i}
    \addConstraint{S^{AUX}_{t-1, i} + M^{out}_{t-1, i}}{\geq S^{AUX}_{t, i}}{\forall t \geq 2 ~~\forall i}
    \addConstraint{S^{AUX}_{t, i}}{\geq M^{in}_{t, i}}{\forall k \in K ~~\forall t \geq 2 ~~\forall i}
    \addConstraint{S^{RAM}_{t, i}}{\geq M^{out}_{t, i}}{\forall k \in K ~~\forall t \geq 2 ~~\forall i}
    \addConstraint{U^{RAM}_{t,i}}{\leq \mu_{RAM}}{\forall t \in V ~~\forall i \in V}
    \addConstraint{\sum_T [R \Psi_{compute}]_T}{\leq \mu_{deadline}}{}
    \addConstraint{S_{1, i}}{=0}{\forall i \in V}
    \addConstraint{\textstyle R_{v,v}}{= 1}{\forall v \in {V}}
    \addConstraint{R, S_{SD}, S_{RAM}, M_{in}, M_{out}}{\in \{0, 1\}^{T \times T}}
\end{argmini}
\caption{\textbf{\poet{} optimizer}: The complete memory constrained MILP with $O(|V||E|)$ variables and constraints. The definition of $U$ (not listed) is from~\citet{checkmate}, Equations 2 and 3.}
\label{algo:full_poet_formulation}
\end{algorithm*}
\textbf{Assumptions:}~~We assume operations execute sequentially on edge devices without inter-operator parallelism. Moreover, we assume parameters and gradients are stored in a contiguous memory region without paging. Unlike prior work in rematerialization~\cite{chen2016b,dtr}, we do not limit rematerialization to occur once. 
We assume auxiliary storage (e.g., flash/ SD card) is available. 
However, if auxiliary storage is not available, the \poet{} optimizer will fall back to only performing rematerialization.

\subsection{
Optimal Rematerialization}

Following the design of Checkmate~\cite{checkmate}, we introduce the formulation of the rematerialization problem.
Given a directed acyclic dataflow graph $G = (V, E)$ with $n$ nodes, a topological ordering $\{v_1, \ldots, v_n\}$ is computed which constrains execution to that order of instructions. Two key decision variables are introduced: (1) $R \in \{0,1\}^{n \times n}$ where $r_{t,i}$ represents the decision to (re)materialize an operation $v_i$ at timestep $t$ and (2) $S \in \{0,1\}^{n \times n}$ where $s_{t,i}$ represents whether the result of an operation $v_i$ is resident in memory at timestep $t$.

From the rematerialzation matrix $R$, and the storage matrix $S$, we define a series of constraints to maintain graph dependencies. All arguments for an operation $j$ must be resident in memory prior to running that operation, yielding constraint $R_{t,i} + S_{t,i} \geq R_{t,j}~~\forall (i,j) \in E~~\forall t \in \{1,\ldots,n\}$. Similarly, the result of an operation is only resident in memory in one of the two cases: a) if it was already resident in memory before, or b) if it was (re)materialized ($S_{t,i} \leq S_{t-1,i} + R_{t-1,i}~~\forall i \in V~~\forall t \in \{1,\ldots,n\}$).

To adhere to the strict constraints on the peak memory used during training, an intermediate variable $U \in \mathbb{R}^{n \times n}$ is defined. $U_{t,i}$ is the total memory used by the system during training at timestep $t$ when evaluating operation $i$. By bounding the maximum value of $U_{t,i}~~\forall i \in V~~\forall t \in \{1,\ldots,n\}$ to the user-specified memory limit $\mu_{RAM}$, we limit the total memory consumption during training.

\subsection{Optimal integrated paging and rematerialization}
While rematerialization can provide significant memory savings, it introduces significant energy consumption overheads from duplicate recomputations. Similarly, paging if done wrong will result in a wasteful shuffling of data between memories. Here, we formalize a joint search space for rematerialization and paging to enable the discovery of the energy-optimal hybrid schedule.

Like rematerialization, the discovery of the optimal paging schedule is a challenging combinatorial search problem. However, we find that independently solving for paging first, and then solving for rematerialization will not produce globally optimal solutions. As an example, consider a graph where the output depends on the result of two operations $v_1$ and $v_2$ where both nodes have equivalent memory costs but $v_2$ is cheaper to evaluate. A paging strategy may evict $v2$ which would force rematerialization to recompute the more expensive $v_1$ rather than $v_2$.

We represent a schedule as a series of nodes that are either being saved $S^{RAM}$, (re)computed $R$ or paged from secondary storage $S^{AUX}$. To model when a node is copied from secondary storage to RAM, we introduce a variable $M^{in} \in \{0, 1\}^{n \times n}$ where $M^{in}_{t,i}$ represents paging a tensor from secondary storage to RAM between timesteps $t-1$ and $t$. Similarly, we model page-out with $M^{out}$.

We now present the intuition behind adding the following constraints to the optimization problem in order to search over optimal schedules for paging and rematerialization:
\begin{itemize}
    \item[1c] For $S_{t,i}^{RAM}$ to be in memory at time-step $t$, either compute $R_{t,i}$ at timestep $t$, or retain $S_{t,i}^{RAM}$ in memory from the previous timestep $t-1$, or page-in if $S_{t-1,i}^{RAM}$ is resident on flash (at $t-1$).
    \item[1d] Each node $i$ can reside on flash $S_{t,i}^{AUX}$, either if it resided on flash at timestep $t-1$ ($S_{t-1,i}^{AUX}$), or it was paged out at time-step $t-1$ ($M_{t-1,i}^{out}$).
    \item[1e] To page-in $M_{t,i}^{in}$ at time-step $t$, it has to be resident on flash $S_{t,i}^{AUX}$ at timestep $t$.
    \item[1f] Each node $i$ can reside in memory $S_{t,i}^{RAM}$ at timestep $t$, only if it was paged out of flash ($M_{t,i}^{out}$).
\end{itemize}
Algorithm~\ref{algo:full_poet_formulation} defines the complete optimization problem.

\subsection{Expressing an energy consumption objective}
If we only consider rematerialization, then minimizing runtime will generally correlate with decreased energy usage. However, this is no longer true when considering paging; paging can be more energy-efficient than rematerializing a compute-intensive operation. To address this, we introduce a new objective function to the optimization problem that minimizes the combined energy consumption due to computation, page-in, and page-out.

When paging occurs on an edge device, the vast majority of energy consumed is due to powering-on the flash/SD block device. As this power is in addition to any power the CPU is consuming, the total power consumption is a linear combination of paging and CPU energy. We precompute each of these values, generally as the integral of the power of active components of the edge device integrated over the runtime of the operation. $\Phi_{compute}$, $\Phi_{pagein}$ and $\Phi_{pageout}$ represent the energy consumed for each node for computing, paging in, and paging out respectively.

Therefore, the new objective function combining paging and rematerialization energy usage is:
\begin{equation}
\sum_T \left[R \Phi_{compute} + M_{in} \Phi_{pagein} + M_{out} \Phi_{pageout}\right]_T
\end{equation}

\subsection{Ensuring minimum training throughput}

If we attempt to find the minimum energy schedule subject to only a memory constraint, the \poet{} solver may select solutions with poor end-to-end training throughput. Ideally, training should occur in the downtime between interactive workloads on an edge device. To ensure this, we introduce a new constraint to the optimization problem that ensures schedules meet a minimum training throughput threshold. This constraint effectively trades off between energy consumption and training throughput.

To enforce a particular throughput, we compute a latency target. Via profiling, we capture $\Psi_{compute}$ denoting the runtime of each operation. We then constrain total runtime with the constraint:
\begin{equation}
\sum_T [R \Psi_{compute}]_T \leq \mu_{deadline}
\end{equation}

\subsection{Paging latency hiding via transfer planner}

\poet{} outputs the DAG schedule in terms of which nodes of the graph ($k$) to  rematerialize, and which to page-in ($M^{in}_{t,k}$) or page-out ($M^{out}_{t,k}$) at each time-step ($t$). Our Algorithm~\ref{alg:poet} takes the ILP solves to generate and dictate the strategy that determines which tensors are resident-in-memory ($S^{aux}_{t,k}$) at a fine-grained (operator) level.

We factor in the latency introduced by paging. As described in Section~\ref{sec:experiments}, \poet{} is hardware-aware by profiling the latency per platform for paging activations to secondary storage. Fine-grained profiling helps in fine-tuning when to start paging, such that the activation tensors arrive just-in-time. We then modify the page-in ($M^{in}_{t,k}$) and the page-out ($M^{out}_{t,k}$) schedule to ensure there is no contention for the memory bus as the tensors are paged-in just-in-time. For example, if ($M^{in}_{t,i}$) can contend with ($M^{in}_{t,j}$), then we schedule one of them to page-in at an earlier time ($M^{in}_{t-1,i}$) and update the in-memory schedule ($S^{aux}_{t-1,i}$) to account for the earlier paging-in. While this ensures the activations are paged in just-in-time, in parallel, ($R_{t',j}$) informs the PyTorch DAG scheduler to deallocate the tensors that we have chosen to rematerialize ($S^{aux}_{t,k}$) at a future timestep ($t'$). 

\begin{algorithm}[t]
\label{algorithm:poet}
	\DontPrintSemicolon
	\KwIn{Graph $G= (V,E)$, schedule $R, M_{in}, M_{out}$}
	\For{t=1,..,$|V|$}{
		\For{k=1,..,$|V|$}{
		    \If{$M^{in}_{t}$}{
		        add \texttt{\%r = pagein $v_k$} to $P$
	        }
			\If{$R_{t,k}$}{
				add \texttt{\%r = compute $v_k$} to $P$
			}
		    \If{$M^{out}_{t,k}$}{
		        add \texttt{\%r = pageout $v_k$} to $P$
	        }
	        \For{$i \in \text{DEPS}[k] \cup \{k\}$}{
    	        \If{$M^{out}_{t,k} \lor \text{FREE}_{t,i,k}$}{
    		        add \texttt{deallocate \%r} to $P$
    	        }
	        }
		}
	}
	\KwOut{execution plan $P=(v_1, .., v_n)$}
	\caption{Training Graph Execution Plan}
	\label{alg:poet}
\end{algorithm}

\section{Evaluation}

\begin{table}[t]
\centering
\begin{tabular}{llllc}
\toprule
\textbf{Device} & \textbf{Clock}  & \textbf{RAM} & \textbf{FPU?}  \\ \midrule
M0 (MKR1000)     & 48 MHz & 32 KB   & $\times$ \\
M4 (nrf52840)    & 64 MHz & 256 KB & $\surd$  \\
A72 (RPi-4B+).   & 1.5 GHz & 2 GB & $\surd$ \\
A57 (Jetson TX2) & 2 GHz & 8 GB & $\surd$ \\ \bottomrule
\end{tabular}
\caption{We evaluate a wide variety of battery-powered edge devices. All devices have at least 32GB of flash memory via an SD card or flash to enable paging of activations or tensors. FPU is floating-point unit.}
\label{table:device-spec}
\end{table}

In our evaluation of \poet{} we seek to answer three key questions. First, how much energy consumption does \poet{} reduce across different models and platforms? Second, how does \poet{} benefit from the hybrid paging and rematerialization strategy? Lastly, how does \poet{} adapt to different runtime budgets?

\subsection{Experimental setup}
\label{sec:experiments}
We evaluate \poet{} on four distinct hardware devices listed in Table~\ref{table:device-spec}: the ARM Cortex M0 class MKR1000, ARM Cortex M4F class nrf52840, A72 class Raspberry Pi 4B+, and Nvidia Jetson TX2. 
\poet{} is fully hardware-aware and relies on fine-grained profiling. For example, on the Jetson-TX2 hardware we profile each operator along with its variations in dimensionality (e.g., \texttt{conv2d} with varying kernel-sizes, strides, padding, etc.) These fine-grained time, energy, and memory profiles then inform \poet{} about the exact specifications. 
These devices test a diverse set of memory, compute, and power configurations. As \poet{} is hardware and energy-aware, it takes device-specific characteristics into account.

We evaluate \poet{} on VGG16~\cite{vgg} and ResNet-18~\cite{resnet} trained on the CIFAR-10 dataset as well as BERT~\cite{bert}.
In all of our baselines, we limit all MILP solves to no more than
10 min on commodity CPUs. Our experiments are with a batch-size of 1. 
We compare \poet{} to work PyTorch's default scheduler, \citet{chen2016b}, \citet{griewank}, DTR~\cite{dtr}, and Checkmate~\cite{checkmate}.

\textbf{Hyperparameters:}~\poet{} only decides on the optimal scheduling of nodes in the training graph and does $not$ change the training routine (learning rate, optimizer, etc.). Hence, our system is robust to hyper-parameters. 

Sensitivity to Batch-size: POET is mathematically preserving and can be easily scaled to arbitrary batch sizes without loss of generality. Of course, this is conditioned on the underlying device's memory capacity. It is possible that as  batch size varies, the underlying operator implementation might change. POET, with its fine-grained profiling is robust to these changes and transparently adapts to  artifacts.

\subsection{How much energy consumption does \poet{} reduce across models and platforms?}

\label{sec:method:transfer_planner}

\begin{figure*}[t]
    \centering
    \includegraphics[width=\textwidth]{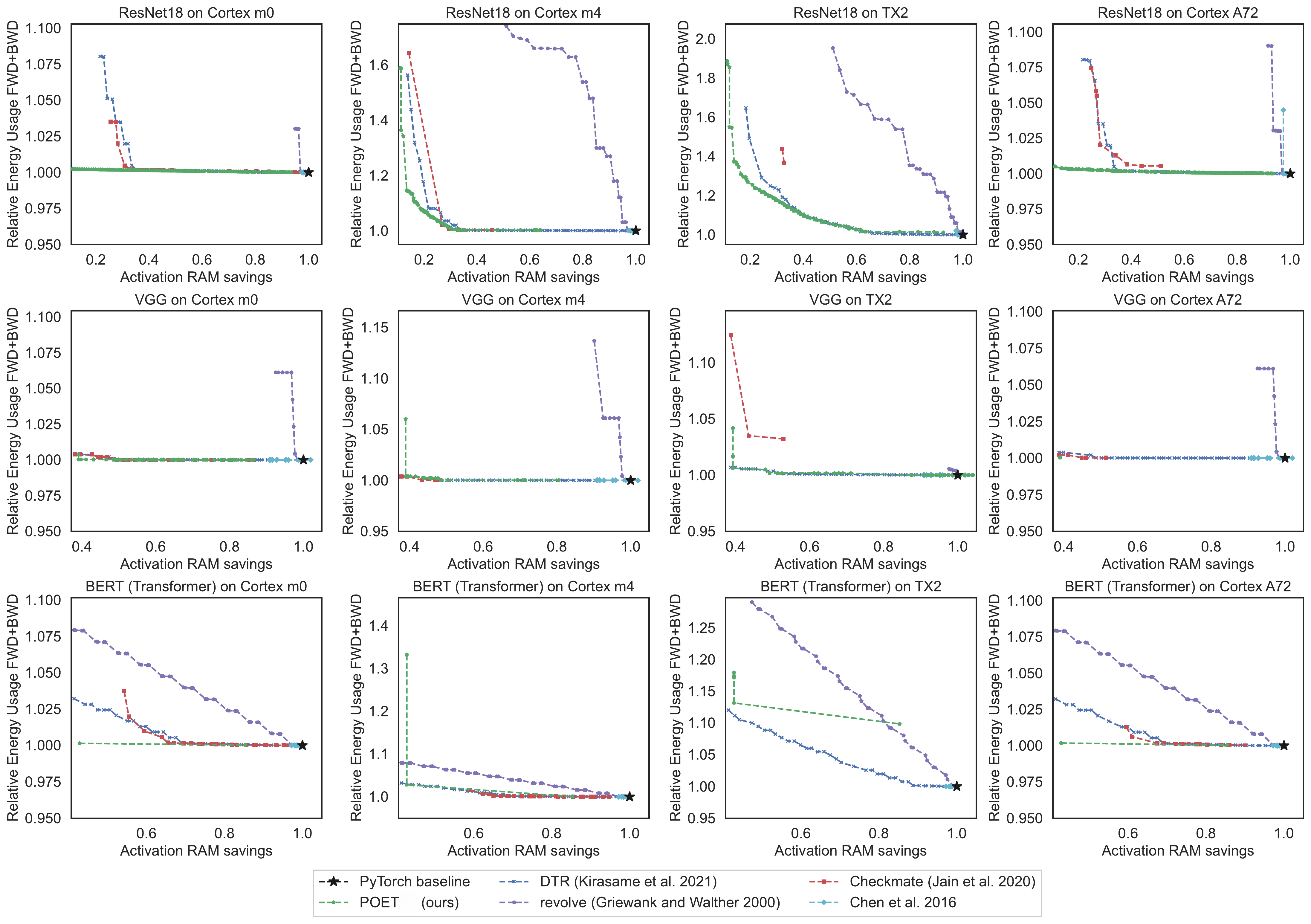}
    \caption{
\textbf{\poet{} consumes less energy across diverse models and devices}:~~We profile the energy usage of each method relative to a full-memory configuration as the device's available memory capacity shrinks. For ResNet-18 (top row), VGG (middle row) and BERT (bottom), \poet{} outperforms competitive methods in most configurations. When training ResNet-18 on the TX2, \poet{} consumes up to 35\% less energy than DTR while discovering solutions at tighter memory budgets.
}
    \label{fig:pareto_flops}
\end{figure*}

\begin{figure*}[t]
    \centering
    \includegraphics[width=\linewidth]{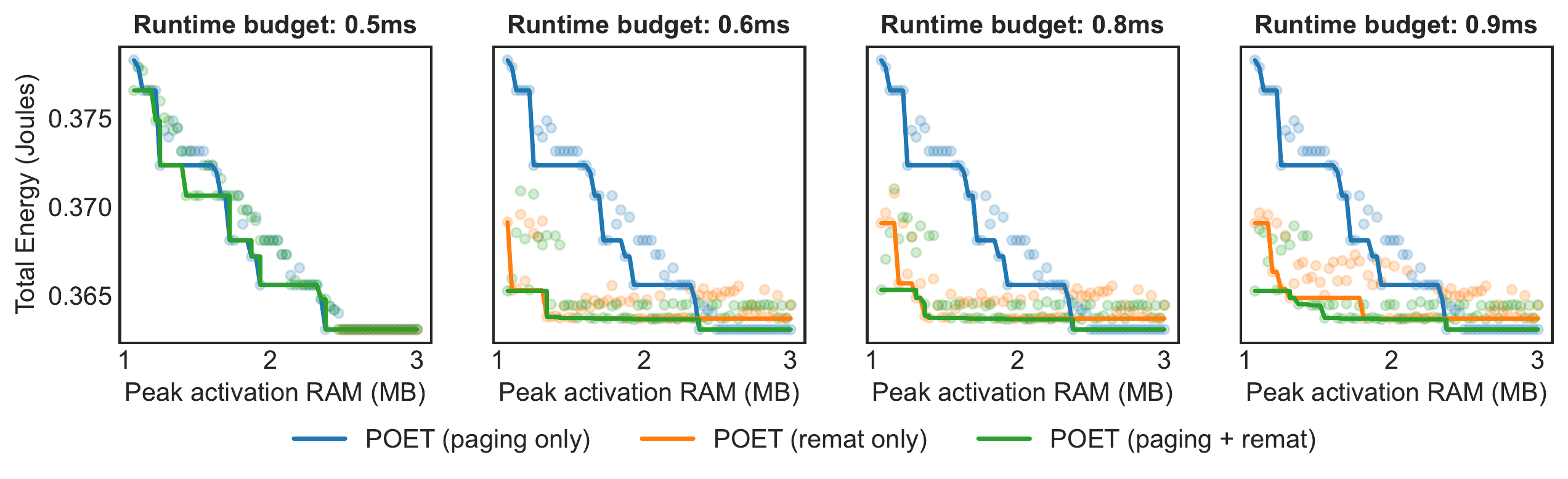}
    \caption{\textbf{Both rematerialization and paging are necessary for low-energy schedules with limited memory}:~~We compare ablations of \poet{} on VGG for CIFAR-10 and find that both rematerialization and paging are required to achieve low-energy solutions at limited memory budgets across all runtime constraint values.
    }
    \label{fig:timing}
\end{figure*}

\begin{figure}[t]
    \centering
     \includegraphics[width=0.85\columnwidth]{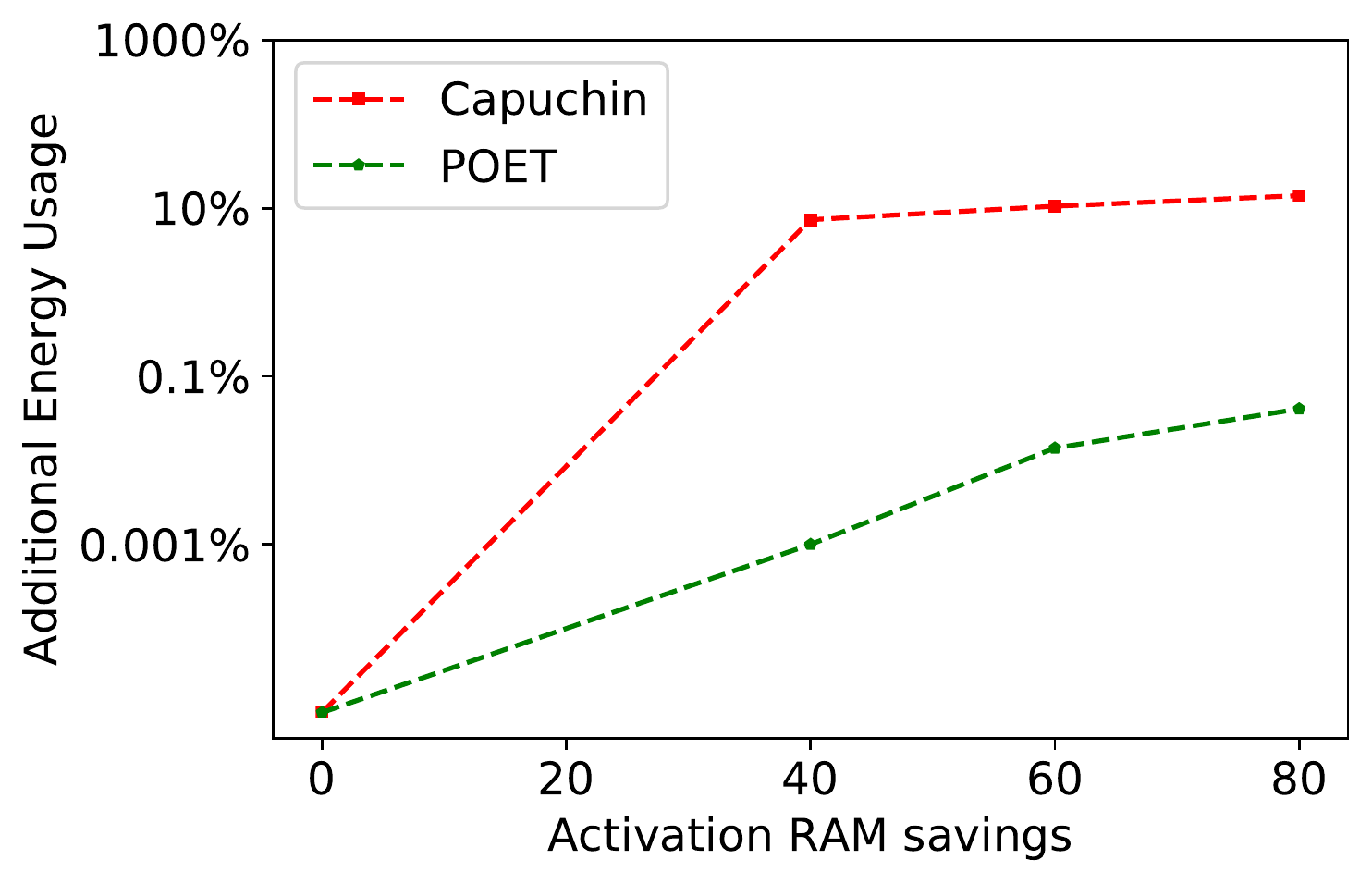}
    \caption{\textbf{Optimal integrated rematerialization and paging outperforms Capuchin (log scale)}:~~\poet{} incurs 73\% to 140\% less energy overhead relative to a full-memory baseline by rematerializing earlier alongside paging. Capuchin strongly prefers paging before falling-back on rematerializing activations which makes it  sub-optimal.
    }
    \label{fig:capuchin}
\end{figure}

\begin{table}[t]
\centering
\begin{tabular}{@{}rcc@{}}
\toprule
          & \multicolumn{2}{c}{\textbf{ResNet-18 Training}} \\ 
          & \poet{}            & POFO (Beaumont et al. 2021)     \\ \midrule
Memory & 285,873 kB             & 311,808 kB \\
Runtime & 82.36 ms &  94.79 ms \\ \bottomrule
\end{tabular}
\caption{\poet's MILP formulation lowers peak memory consumption by
8.3\% and improves throughput by 13\% compared to POFO \cite{beaumont2021} on Nvidia's Jetson TX2 edge device}
\label{table:beaumont}
\end{table}

Figure~\ref{fig:pareto_flops} shows the energy consumed for a single epoch of training. Each column represents a unique hardware platform as defined in Table 2. We notice that across all platforms, \poet{} generates the most energy-optimal (Y-axis) schedule all the while reducing the peak memory consumed (X-axis) and adhering to the timing budget.

For the BERT model on the Cortex M4 and the TX2 platform, we noticed an interesting behavior: our ILP solves time-out. This is because we limit all solves to no more than 10 min. With a longer ILP solve budget ($<$30 min), \poet{} can predictably find more optimal solutions. Further, notice that a) \poet{} has an additional timing budget which none of the other baselines do, and b) all of our baselines are already mature. Checkmate~\cite{checkmate} is provably optimal for rematerialization, while DTR~\cite{dtr} closely approximated Checkmate. Furthermore, \poet{} tried to solve a much ``harder'' problem as its search space with rematerialization and paging together is larger.

\subsection{How does \poet{} benefit from integrated rematerialization and paging?}

We compare our joint optimal paging and rematerialization schedule with Capuchin which optimizes each with a heuristic. Capuchin will effectively page until no longer feasible and only then will it begin to rematerialize. Instead, \poet{} begins rematerializing cheap operations like ReLU much earlier which yields considerable energy savings (up to 141\% lower overhead).

In Figure~\ref{fig:capuchin}, we benchmark \poet{} and Capuchin when training ResNet-18 on the A72. As the RAM budget decreases (to the right), Capuchin consumes 73\% to 141\% more energy than a baseline with full memory. In comparison, \poet{} incurs less than a 1\% energy overhead. 
This trend holds for all architectures and platforms we tested.

In Table~\ref{table:beaumont} we benchmark \poet{} and POFO when training ResNet-18 on Nvidia's Jetson TX2. We find that \poet{} finds an integrated rematerialization and paging schedule that lowers \textit{peak memory consumption by 8.3\% } and \textit{improves throughput by 13\%}. This showcases the benefit of \poet{}'s Mixed-integer linear programming (MILP) solver, which is able to optimize over a much larger search-space. While POFO only supports linear models, \poet{} generalizes to non-linear models as demonstrated in Fig~\ref{fig:pareto_flops}.

\subsection{How does \poet{} adapt to varying runtimes?} 
Figure~\ref{fig:timing} highlights the benefit of the integrated  strategies that \poet{} adopts across different timing constraints. The run-time budget refers to the total time available for one epoch of training na\"ively (without paging or rematerialization). For each of the runtimes, we plot the total energy consumed if we were to restrict to either of a) paging or b) rematerialization only, and the c) integrated solution.

We find that rematerialization is energy-optimal compared to paging at higher (looser) timing budgets. This is reflected in the POET (paging+remat) green curve, closely tracking the POET (remat only) yellow curve at runtime budgets of $0.6$ - $0.9$ ms. However, at lower runtime budget ($0.5$ ms), paging is preferable as rematerialization strategies become infeasible. This is because, rematerialization is a compute intensive serial operation, however, our Algorithm~\ref{alg:poet} benefits from the ability to hide paging latencies by pipelining (see Section~\ref{sec:method:transfer_planner}) to realize the tighter deadline bounds.
POET's optimal, integrated solution consumes up to ${40\%}$ lower energy compared to paging or rematerialization only solutions.

\section{Conclusion}
Enabling large models to be trained on edge devices is important due to privacy constraints as well as offline operation.
Edge devices deployed in the real-world are powered by tiny microcontrollers that are low-powered, and have limited memory (e.g. $32$ KB).
The low-power and limited memory, coupled with tight timing constraints imposed by real-time systems makes training on the edge challenging.

Our novel mixed-integer linear programming based Power Optimal Edge Training (\poet{}) algorithm enables training on tiny chips with memory as low as 32 KB. Given a {memory budget} and a {timing constraint}, \poet{} finds the most {energy optimal} schedule to train the model by choosing to either rematerialize or page the tensors to secondary storage.

Across a diverse set of models and devices, we discover low-power training schedules at less memory than baselines. \poet{} enables new applications for privacy-preserving personalization of large models like BERT on tiny devices at the edge for the first time. Future directions include integrating activation compression as well as expanding \poet{}'s search space to paging parameters.

\section*{Acknowledgement}

We 
thank  Prateek Jain, Charles Packer, Daniel Rothchild, Alex Smola, Pete Warden, and the anonymous reviewers whose insightful 
comments, and feedback helped improve the paper.
This research is supported by a NSF CISE Expeditions
Award CCF-1730628, and gifts from Amazon Web Services, Ant Group, Ericsson, Facebook, Futurewei, Google,
Intel, Microsoft, Scotiabank, and VMware. 
This work was supported in part by the CONIX Research Center,
one of six centers in JUMP, a Semiconductor Research Corporation
(SRC) program sponsored by DARPA.

\balance
\clearpage
\bibliography{poet,paras_mendeley}
\bibliographystyle{icml2022}
\end{document}